\documentclass[runningheads]{llncs}
\usepackage{graphicx}
\usepackage{amsmath}
\usepackage{adjustbox}
\usepackage{algorithm}
\usepackage{algpseudocode}
\usepackage{algpseudocode}
\usepackage{subcaption}
\usepackage{booktabs} 
\usepackage{longtable} 
\usepackage{comment}
\usepackage{xcolor}
\begin{document}

\title{Intelligent Multi-Document Summarisation for Extracting Insights on Racial Inequalities from Maternity Incident Investigation Reports}
\titlerunning{Intelligent Multi-Document Summarisation}
%

\author{Georgina Cosma\inst{1} \and
Mohit Kumar\inst{1} \and
Patrick Waterson\inst{2} \and Gyuchan Thomas Jun\inst{2}\and Jonathan Back \inst{3}}

\authorrunning{G. Cosma et al.}
%
\institute{Department of Computer Science, School of Science, Loughborough University, UK \and
School of Design and Creating Arts, Loughborough University, UK \and
Health Services Safety Investigations Body (HSSIB), United Kingdom
\email{g.cosma@lboro.ac.uk}\\
}
\maketitle              

\begin{abstract}
In healthcare, thousands of safety incidents occur every year, but learning from these incidents is not effectively aggregated. Analysing incident reports using AI could uncover critical insights to prevent harm by identifying recurring patterns and contributing factors. To aggregate and extract valuable information, natural language processing (NLP) and machine learning techniques can be employed to summarise and mine unstructured data, potentially surfacing systemic issues and priority areas for improvement. This paper presents I-SIRch:CS, a framework designed to facilitate the aggregation and analysis of safety incident reports while ensuring traceability throughout the process. The framework integrates concept annotation using the Safety Intelligence Research (SIRch) taxonomy with clustering, summarisation, and analysis capabilities. Utilising a dataset of 188 anonymised maternity investigation reports annotated with 27 SIRch human factors concepts, I-SIRch:CS groups the annotated sentences into clusters using sentence embeddings and k-means clustering, maintaining traceability via file and sentence IDs. Summaries are generated for each cluster using offline state-of-the-art abstractive summarisation models (BART, DistilBART, T5), which are evaluated and compared using metrics assessing summary quality attributes. The generated summaries are linked back to the original file and sentence IDs, ensuring traceability and allowing for verification of the summarised information. Results demonstrate BART's strengths in creating informative and concise summaries.

\end{abstract}

\section{Introduction}
In recent years, deep learning (DL) and natural language processing (NLP) models have demonstrated immense potential for automatically analysing and summarising textual data across application areas. Within the healthcare domain, key use cases include generating concise summaries of electronic health records (EHRs) to accelerate access to patient histories, extracting key findings from the latest medical literature to keep professionals informed of advancements, and simplifying health education content into more readable materials for patient consumption. However, applying DL in sensitive domains like healthcare also introduces confidentiality, privacy, and ethical considerations which demand accurate yet traceable model behaviours that respect patient data sensitivity. To address the ethical considerations and data privacy requirements when applying deep learning models to sensitive healthcare data, it is crucial to ensure that any datasets used are fully anonymised before analysis or modelling, with all personally identifiable information removed from the outset.

Our prior work saw the development of the Intelligence-Safety Intelligence Research (I-SIRch) framework/tool \cite{Singh2023I-SIRch}, which utilises multi-label text annotation with the Safety Intelligence Research (SIRch) human factors taxonomy to systematically categorise contributing factors in adverse maternal care incidents. Trained on a corpus of real (anonymised) and tested on real and synthetically generated maternity investigation reports from UK cases, I-SIRch demonstrated robust performance across numerous evaluation metrics, including recall, precision, and balanced accuracy \cite{Singh2023I-SIRch}. Our research represented an important step towards utilising advanced NLP techniques to enhance patient safety and address care quality gaps, especially within maternal health across different demographics. This paper aims to advance the I-SIRch framework \cite{Singh2023I-SIRch} by proposing an extended I-SIRch:CS\footnote{https://github.com/gcosma/I-SIRchpapers} system with robust multi-document summarisation capabilities to effectively analyse maternal incident investigation reports across various dimensions, including generating concise yet sufficiently informative summaries, ensuring complete traceability back to source data, and facilitating comparative equity evaluations by constructing separate summaries across ethnic groups to identify any care quality disparities underlying adverse incidents.

\section{Related work}
Models including Long Short-Term Memory (LSTM) networks and Transformer-based architectures have been at the forefront, leveraging their ability to understand context and sequence in texts. Specifically, the use of pretrained models such as BERT (Bidirectional Encoder Representations from Transformers) and its adaptations in the medical domain (e.g., BioBERT, ClinicalBERT) have demonstrated promising results in improving the accuracy and relevance of summaries. For instance, LSTM and Transformer models have been applied to summarise patient Electronic Health Records (EHRs), extracting critical information that can aid in diagnosis, treatment planning, and patient monitoring \cite{Luo2023}, \cite{Zhu2023}, \cite{Patil2023}, \cite{Zelina2023}, \cite{Tsai2022}. These summaries provide a comprehensive view of a patient's medical history, current condition, and potential risks, distilled into a format that is easily accessible to healthcare providers. The introduction of models such as BERT (Bidirectional Encoder Representations from Transformers) \cite{BERT} and GPT (Generative Pretrained Transformer) \cite{GPT} marked a significant milestone in NLP. These models have shown remarkable success in various NLP tasks, including text summarisation. Their ability to capture deep contextual relationships within text makes them particularly suited for summarising complex healthcare documents. One notable advancement is the adaptation of these models for domain-specific tasks. For instance, BioBERT \cite{biobert} and ClinicalBERT \cite{ClinicalBERTMC} have been fine-tuned from BERT for biomedical and clinical text processing, respectively, showing improved performance in tasks such as disease classification and patient information summarisation. Moreover, transformer-based models like BART (Bidirectional and Auto-Regressive Transformers) \cite{BART} and T5 (Text-to-Text Transfer Transformer) \cite{T5} have been utilised for both extractive and abstractive summarisation. BART, in particular, has been effective in generating coherent summaries of medical research articles by rephrasing and condensing the original text. While the effectiveness of these models is well-documented, their computational demands pose challenges, particularly in real-time healthcare applications. This has led to the development of more efficient variants, such as DistilBERT \cite{DistilBERT} and MiniLM \cite{MINILM}, which maintain a balance between performance and computational efficiency. These models enable the deployment of advanced NLP techniques in resource-constrained environments, such as mobile health applications and low-resource organisations.

\section{I-SIRch:CS Framework}
We propose an automated system for summarising text data, capable of clustering sentences by similarity, determining the best number of clusters, and generating concise summaries and keywords for each cluster. This process transforms raw text into structured, actionable insights. We utilised offline generative models to ensure patient confidentiality by anonymising and synthesising data, preserving the integrity of insights while safeguarding personal information. \\
Algorithm (\ref{algCS}) provides the text clustering and summarisation process.
\begin{algorithm}
\caption{I-SIRch: CS Text clustering and summarisation} \label{algCS}
\begin{algorithmic}[1] 
\Statex \textbf{Input:}
\Statex \- Set of sentences $S$ with labels $L$ 
\Statex \- Clustering model $M_c$
\Statex \- Summarisation model $M_s$
\For{each label $l \in L$}
   \State Filter $S$ to get subset $S_l$ with label $l$  
   \State Embed sentences in $S_l$ using $M_c$ getting embeddings $E_l$
   \State Determine optimal clusters $k$ using elbow method on $E_l$ 
   \State Cluster $S_l$ into $k$ clusters $\{C_1, \ldots, C_k\}$ using $k$-means on $E_l$
\EndFor
\For{each cluster $C_i$}
   \State Concatenate sentences in $C_i$ into text $T_i$
   \State Generate summary $S_i$ by applying $M_s$ to $T_i$ 
    \State Generate heading $H_i$ by applying $M_s$ to $T_i$ 
   \State Store sentence IDs used in $S_i$ as $ID_i$
\EndFor
\State \textbf{Outputs:} 
\State Summaries $S_i$ and corresponding Headings $H_i$ for each summary 
\State Sentence IDs $ID_i$ used in each summary
\State Evaluation metrics for each summary
\end{algorithmic}
\end{algorithm}
The algorithm takes as input a set of sentences, denoted mathematically as set $\mathcal{S}$, along with an associated set of labels (a.k.a concepts) $\mathcal{L}$ that categorises the sentences. It also requires two machine learning models - a clustering model $M_c$ and a pretrained summarisation model $M_s$. In the first stage, the algorithm groups the input sentences by their labels (i.e. concepts) in $\mathcal{L}$. For each label $l \in \mathcal{L}$, it extracts the subset of sentences $\mathcal{S}_l$ that have that label value $l$ assigned. Next, it applies the clustering model $M_c$ to embed the sentences in a vector space and determine an optimal number of clusters based on the elbow method. The sentences with the same label $l$ are then clustered into these topics, denoted by $\{C_1, C_2, \dotsc, C_k\}$. The second stage is a summary and evaluation loop over the clusters. For each cluster $C_i$ covering a topic, the sentences are concatenated into a single text excerpt $T_i$. This is passed into the summarisation model $M_s$ to generate a summary text $S_i$. Additional metadata including a heading $H_i$ and the sentence IDs used are also extracted for summary $S_i$. Evaluation metrics are then calculated for summary $S_i$ - including quantification of the diversity, relevance, coverage, coherence, conciseness, and readability. These metrics are appended to a master output list. In the end, the full pipeline outputs: the set of summaries $S_i$ for each topic cluster; the headings $H_i$; sentence IDs $ID_i$ used; and evaluation metrics on the different quality attributes for each summary. The algorithm thus performs multi-document summarisation guided by data labels, with integrated quantitative assessment.

\section{Methodology}
\subsection{Dataset of maternity incident investigation reports} 
The Healthcare Services Safety Investigation Branch (HSSIB) provided a random set of 188 anonymised investigation reports describing adverse maternity incidents. The reports were written between 2019 to 2022. The number of reports for each year is as follows: 4 reports in 2019, 115 reports in 2020, 42 reports in 2021, and 27 reports in 2022. Ethnicity was provided for 76 reports. 

\subsection{Experiment methodology}
\begin{figure}[t]
\centering \includegraphics[width=1\linewidth]{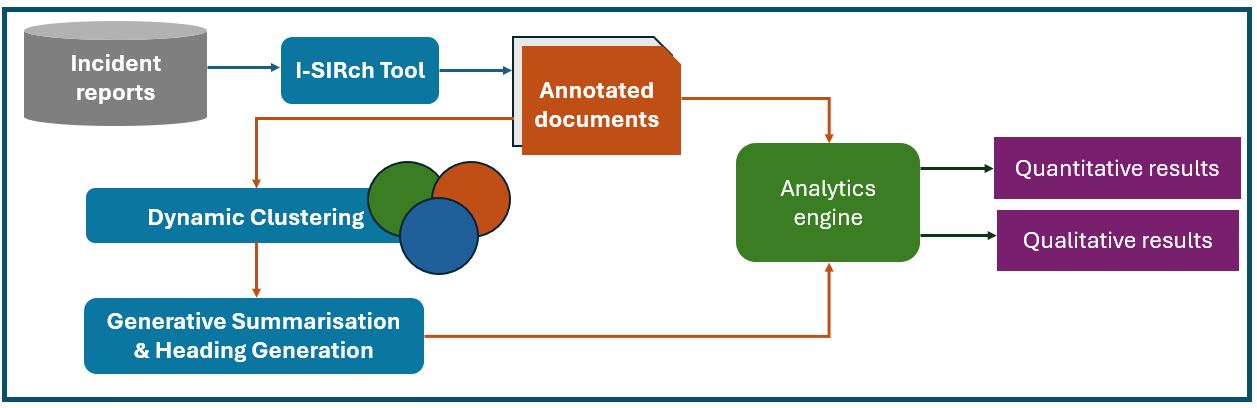}
\caption{I-SIRch:CS pipeline. Shows how the I-SIRch tool has been extended to include clustering and summarisation capabilities analysing intelligence from maternity incidence investigation reports.} \label{framework}
\end{figure}

The I-SIRch:CS framework is presented in Fig.~\ref{framework} and its components are described as follows. \textbf{Healthcare reports repository:} The I-SIRch tool is used for loading, cleaning, and preparing reports. During processing, the tool generates a text file in CSV format that includes File IDs, Sentence IDs, sentences, and concepts derived from the SIRch taxonomy. This phase involves cleaning the files by selecting particular sentences from each report, specifically targeting those with negative connotations, references to physical characteristics, and mentions of medication names associated with dispensing. These selected sentences are annotated according to the SIRch taxonomy and aggregated into a CSV file for subsequent clustering and summarisation tasks.  Table \ref{Labels} shows the list of SIRch concepts with the number of sentences per concept shown in brackets.  \textbf{Dynamic clustering:} Employs the \texttt{all-MiniLM-L6-v2} SentenceTransformer model for sentence embeddings and the elbow method with KMeans for clustering. This process groups text data into semantically similar clusters, maintaining traceability via file and sentence IDs. \textbf{Generative summarisation and heading generation:} Selects from predefined offline models (e.g., BART, T5, DistilBART) for summarisation. Keywords extracted by KeyBERT for headings ensure summaries are informative and traceable to original file and sentence IDs. \textbf{Analytics engine (Qualitative results):} Outputs include the textual summaries and their corresponding topic headings, where each summary is linked to each file and sentence ID. \textbf{Analytics engine (Quantitative results):} Outputs include a detailed CSV report with analysis on summarisation models, clustering, generated headings, summaries, and evaluation metrics, ensuring explainability through file and sentence ID traceability. \textbf{Batch processing and process automation:} The framework's scalable design allows for automated batch processing, enhancing the system's ability to handle large datasets efficiently, with a focus on traceability and organisation.
\begin{table}[ht]
\centering
\caption{ID, concept and number of sentences per concept. Total sentences: 3760.}
\label{Labels}
\begin{tabular}{cp{2in}cp{2in}}
\toprule
\textbf{ID} & \textbf{Concept} & \textbf{Index} & \textbf{Label} \\
\midrule
1 & Acuity (54) & 2 & Antenatal (69)\\
3 & Assessment, investigation, testing, screening (381) & 4 & COVID (142) \\
5 & Care Planning (132) & 6 & Communication factor (477) \\
7 & Decision error (169) & 8 & Dispensing, administering (62) \\
9 & Documentation (168) & 10 & Escalation/referral factor (158) \\
11 & Guidance factor (42) & 12 & Interpretation (197)\\
13 & Language barrier (30) & 14 & Local guidance (88) \\
15 & Monitoring (118) & 16 & National and local guidance (80)\\
17 & National guidance (169) & 18 & Obstetric review (147) \\
19 & Physical characteristics (320) & 20 & Physical layout and Environment (35) \\
21 & Psychological characteristics (54) & 22 & Risk assessment (94)\\
23 & Situation awareness (77) & 24 & Slip or lapse (188) \\
25 & Teamworking (530) & 26 & Technologies and Tools-issues (112) \\
27 & Training and education (40) & & \\
\bottomrule
\end{tabular}
\end{table}

\subsection{Methods}
We employed the following transformer-based architectures that have been fine-tuned for summarisation, and which have demonstrated their capability to address complex NLP challenges, particularly in healthcare. \textbf{Sentence Transformers (`all-MiniLM-L6-v2'):} A compact model that generates semantically meaningful sentence embeddings suitable for semantic search, clustering and summarisation, balancing performance with computational efficiency. \textbf{BART (`facebook/bart-large-cnn'):} A transformer model pretrained for generation tasks, including summarisation. The `bart-large-cnn' variant excels at creating coherent and concise summaries, optimised for tasks requiring detailed summary outputs. \textbf{T5 (`t5-small'):} Adapts the text-to-text approach, treating all NLP tasks as such. The `t5-small' version offers a balance between size, speed, and quality, ideal for diverse summarisation needs. \textbf{DistilBART (`sshleifer /distilbart-cnn-12-6'):} A lighter, faster distilled version of BART, maintaining robust summarisation capabilities. Optimised for efficiency, it provides a viable alternative to its parent model for summarisation tasks. 

\subsection{Evaluation metrics} \label{sec:evalmetrics}
Several metrics have been used to assess different aspects of summarisation quality, including diversity, relevance, coverage, coherence, conciseness, and readability. Each metric provides insights into how well the summarisation process captures and conveys the essential information from the original text. Higher values across these metrics indicate better summarisation model performance in producing summaries that are diverse, relevant, comprehensive, logical, concise and readable.\\
\textbf{Diversity}  measures the variety of vocabulary used in the summary. It is calculated as the ratio of unique words to the total number of words in the summary.
\begin{equation}
\text{Diversity} = \frac{\text{Number of Unique Words}}{\text{Total Number of Words}}
\end{equation}
\textbf{Relevance}  assesses the similarity between the original text and its summary. Cosine similarity between the vector representations of the original text and the summary is often used for this purpose.
\begin{equation}
\text{Relevance} = \cos(\theta) = \frac{\mathbf{A} \cdot \mathbf{B}}{\|\mathbf{A}\| \|\mathbf{B}\|}
\end{equation}
where $\mathbf{A}$ and $\mathbf{B}$ are the vector representations of the original text and the summary, respectively.\\
\textbf{Coverage} evaluates the extent to which the summary captures the key concepts of the original text. It can be quantified by the proportion of original text tokens (or concepts) that appear in the summary.\\
\begin{equation}
\text{Coverage} = \frac{\text{Number of Unique Tokens in Both Summary and Original Text}}{\text{Number of Unique Tokens in Original Text}}
\end{equation}
\textbf{Coherence} measures the logical flow and connection between sentences within the summary. While more challenging to quantify, coherence can be assessed by evaluating sentence embeddings' similarity in sequence.
\begin{equation}
\text{Coherence} = \frac{1}{N-1} \sum_{i=1}^{N-1} \cos(\theta_{i, i+1})
\end{equation}
where $N$ is the number of sentences in the summary, and $\cos(\theta_{i, i+1})$ is the cosine similarity between consecutive sentence embeddings.\\
\textbf{Conciseness} indicates the brevity of the summary. It can be inversely related to the summary's length, encouraging summaries that convey information efficiently.
\begin{equation}
\text{Conciseness} = \frac{1}{\text{Word Count of Summary}}
\end{equation}
\textbf{Readability} measures how easy it is to understand the summary. The Flesch Reading Ease score is a common metric, calculated as follows:
\begin{equation}
\text{Readability} = 206.835 - 1.015 \left( \frac{\text{Total Words}}{\text{Total Sentences}} \right) - 84.6 \left( \frac{\text{Total Syllables}}{\text{Total Words}} \right)
\end{equation}

\section{Results}
The IDs of sentences assigned to each cluster are tracked and used for generating the summary of that cluster. Hence, all sentences grouped into a specific cluster are considered in the summarisation process. However, the actual summary may not incorporate these sentences verbatim. Instead, the summary provides a condensed version that captures the main points from these sentences. An analysis of the summaries is presented below. 
\subsection{Performance comparison of summarisation methods}
\begin{figure}[]
\centering
\begin{subfigure}{.32\textwidth}
  \centering
  \includegraphics[width=\linewidth]{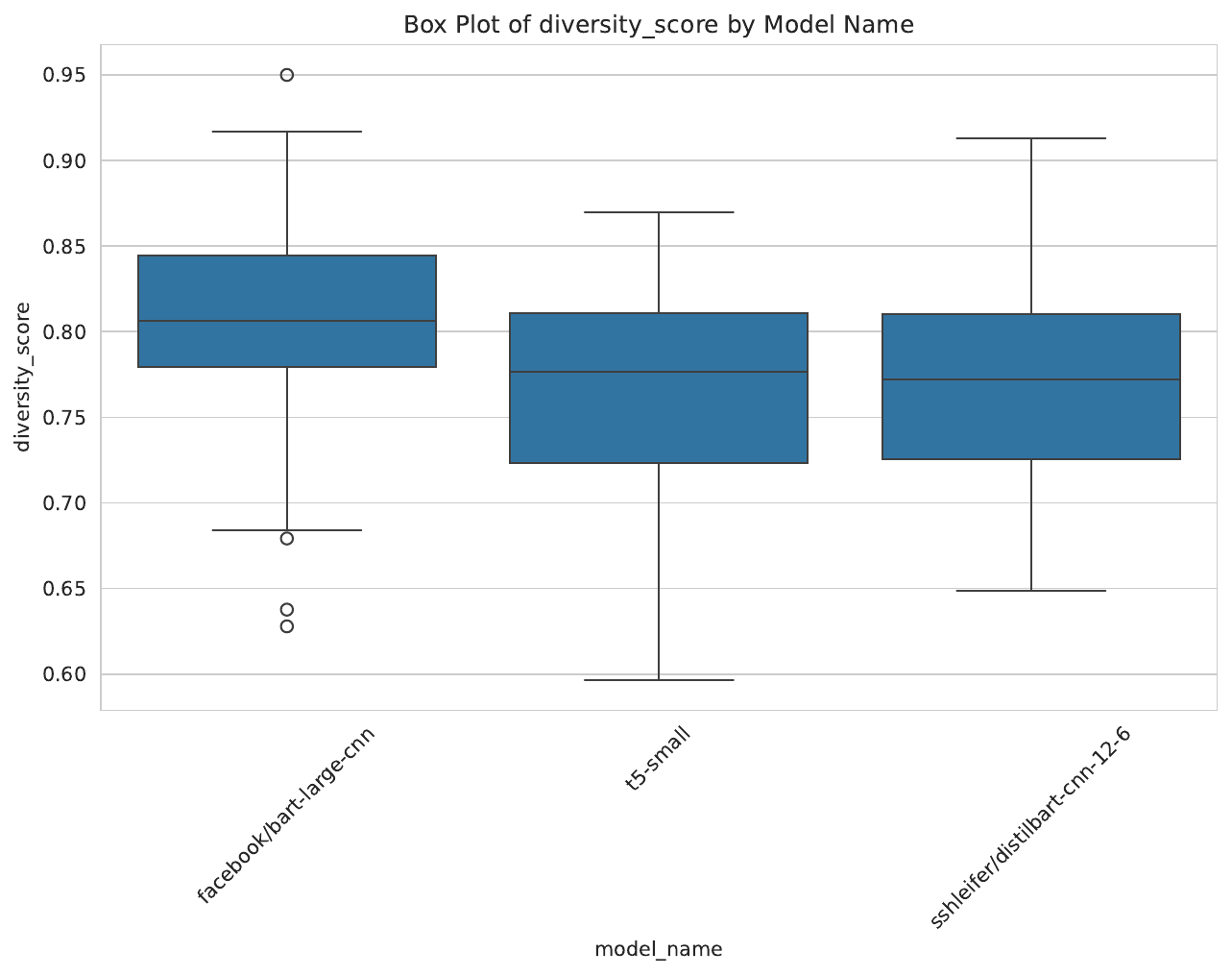}
  \caption{Diversity score boxplot}
  \label{fig:diversity_score}
\end{subfigure}%
\hfill
\begin{subfigure}{.32\textwidth}
  \centering
\includegraphics[width=\linewidth]{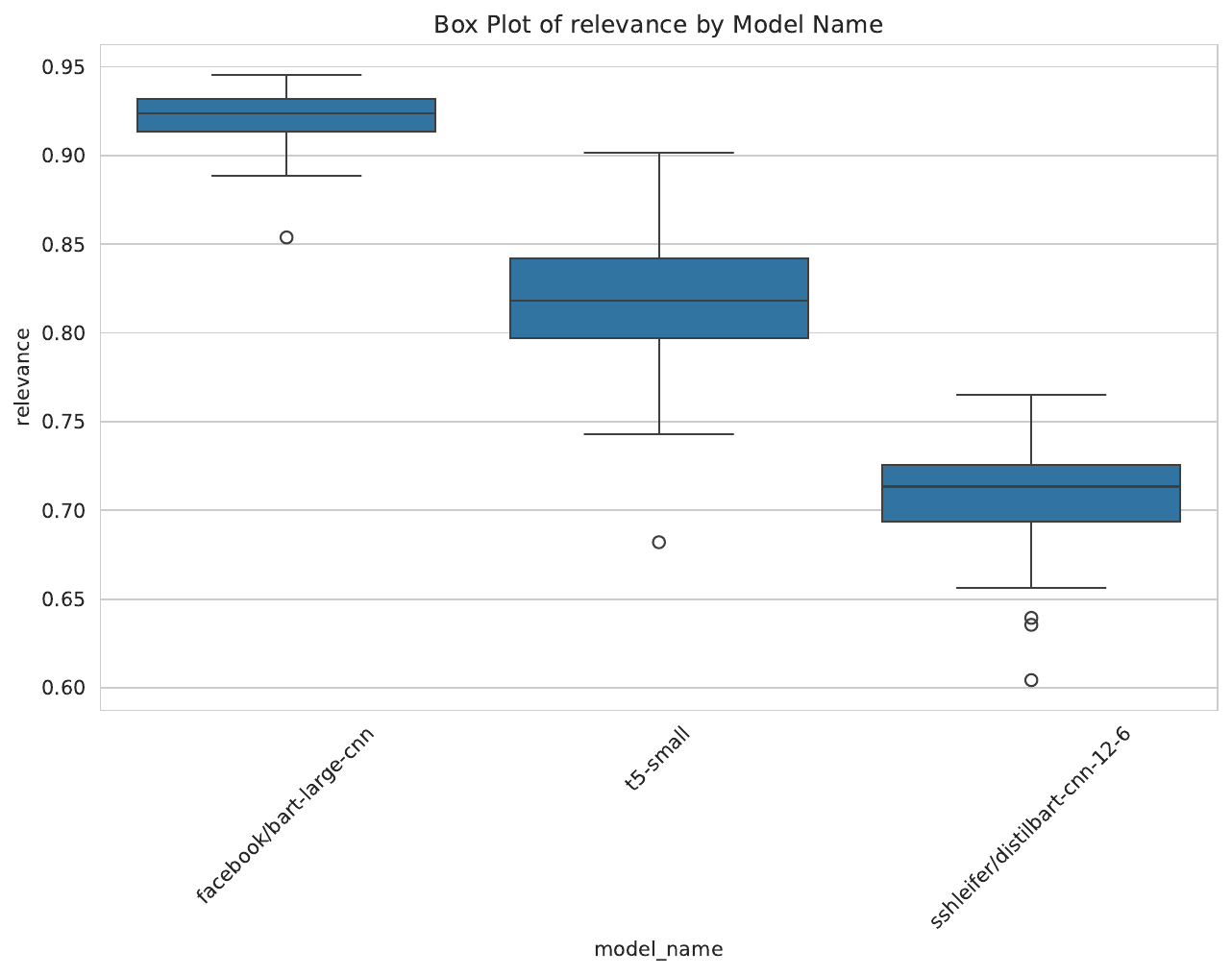}
  \caption{Relevance boxplot}
  \label{fig:relevance}
\end{subfigure}%
\hfill
\begin{subfigure}{.32\textwidth}
  \centering
  \includegraphics[width=\linewidth]{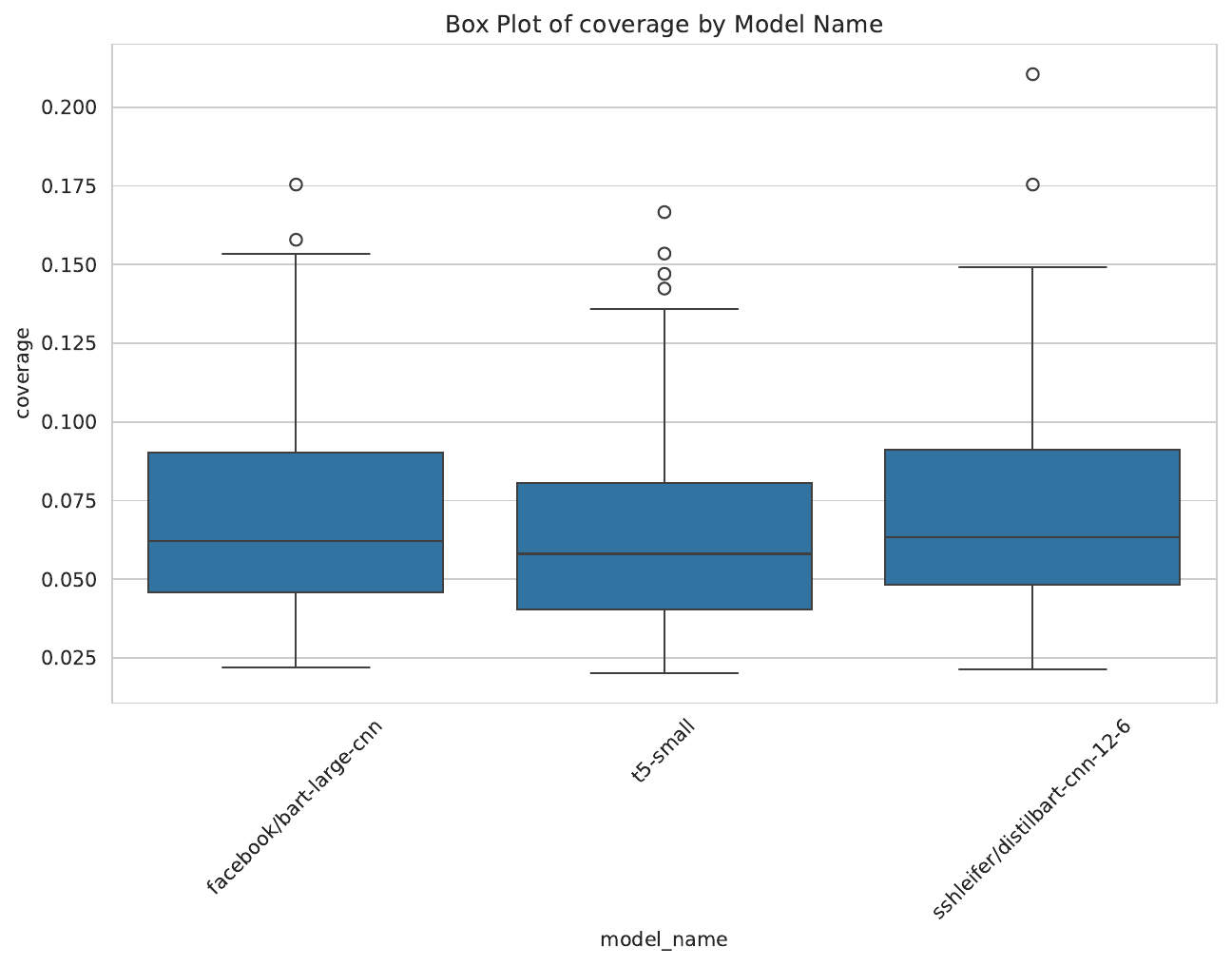}
  \caption{Coverage boxplot}
  \label{fig:coverage}
\end{subfigure}

\begin{subfigure}{.32\textwidth}
  \centering
  \includegraphics[width=\linewidth]{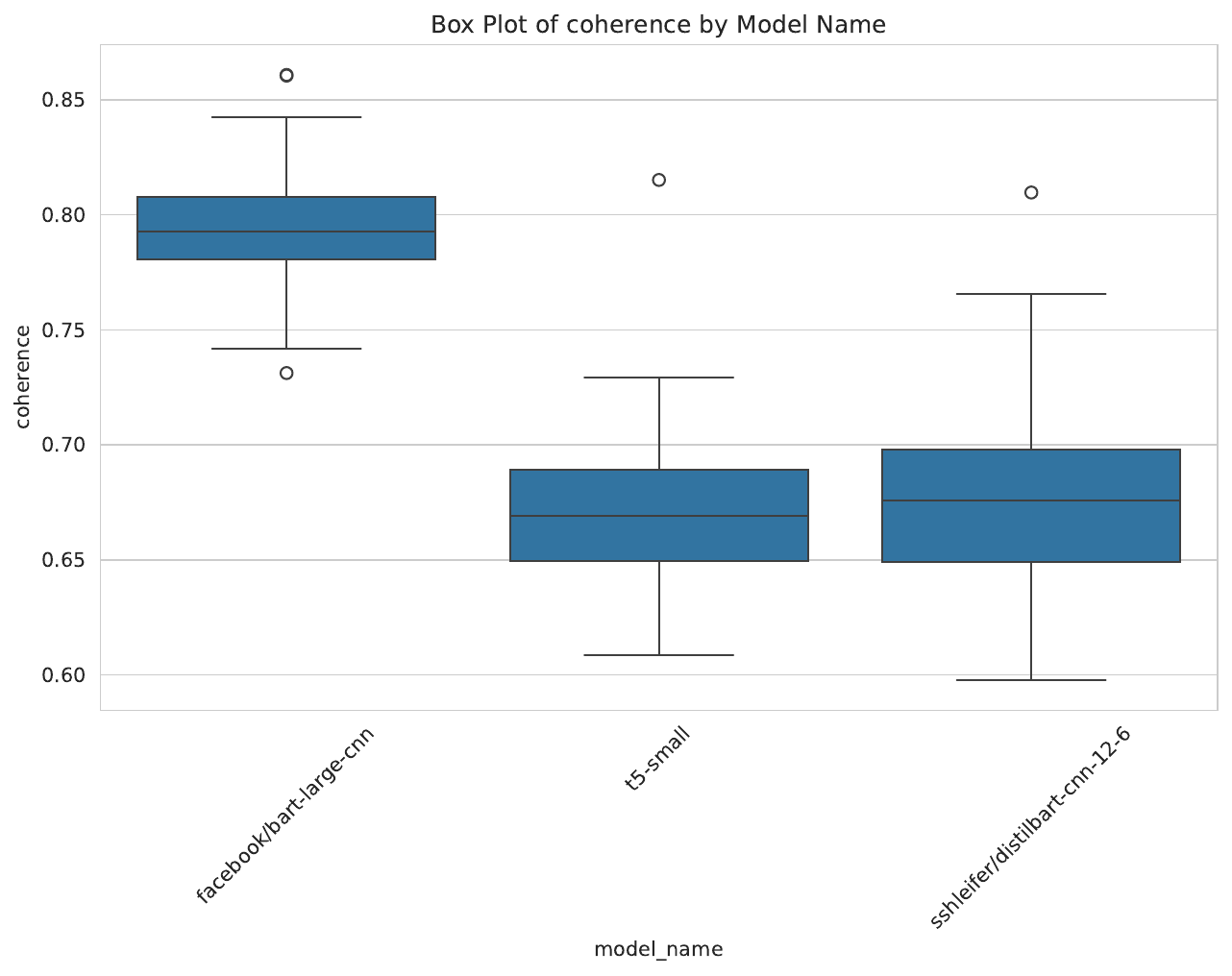}
  \caption{Coherence boxplot}
  \label{fig:coherence}
\end{subfigure}%
\hfill
\begin{subfigure}{.32\textwidth}
  \centering
  \includegraphics[width=\linewidth]{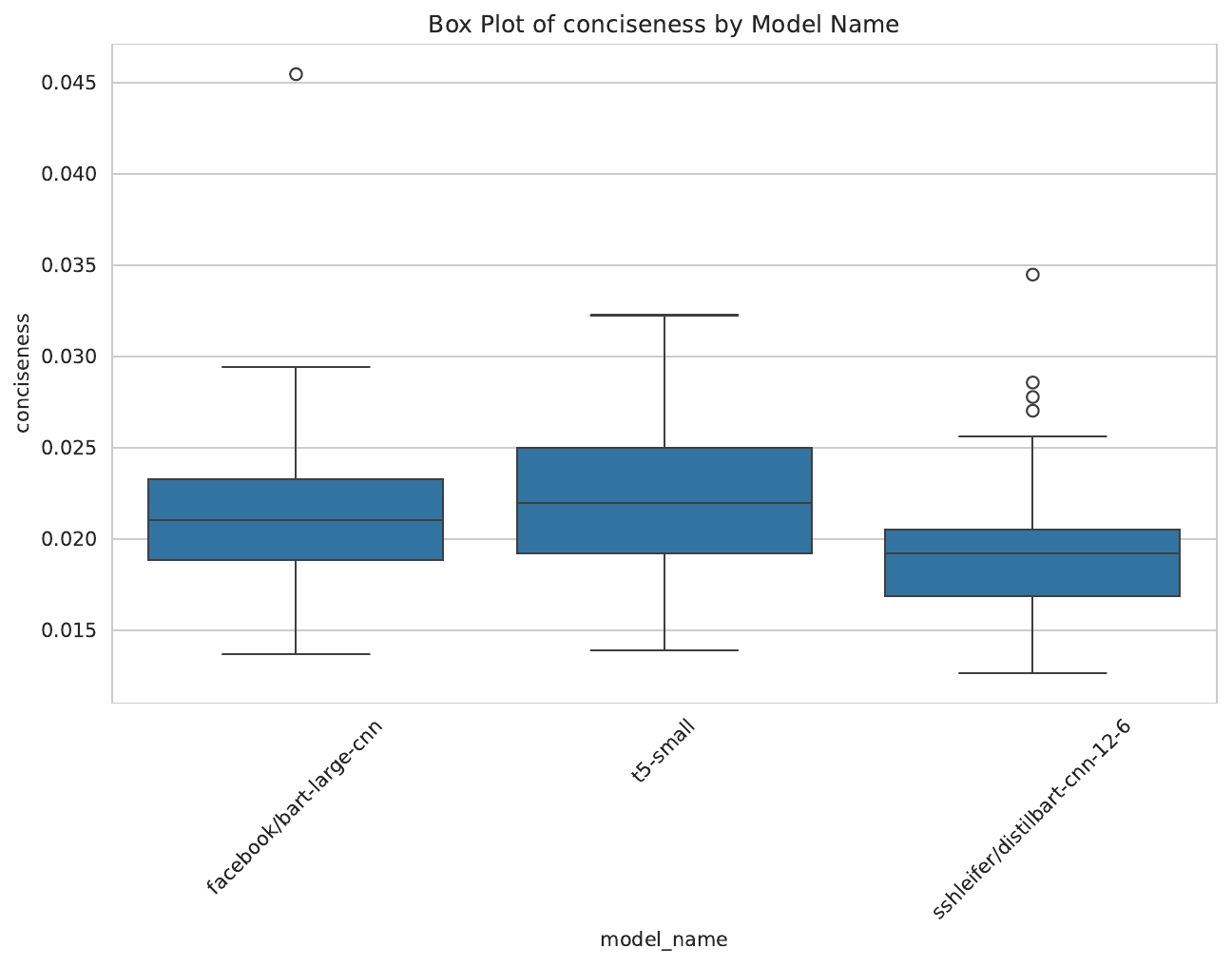}
  \caption{Conciseness boxplot}
  \label{fig:conciseness}
\end{subfigure}%
\hfill
\begin{subfigure}{.32\textwidth}
  \centering
  \includegraphics[width=\linewidth]{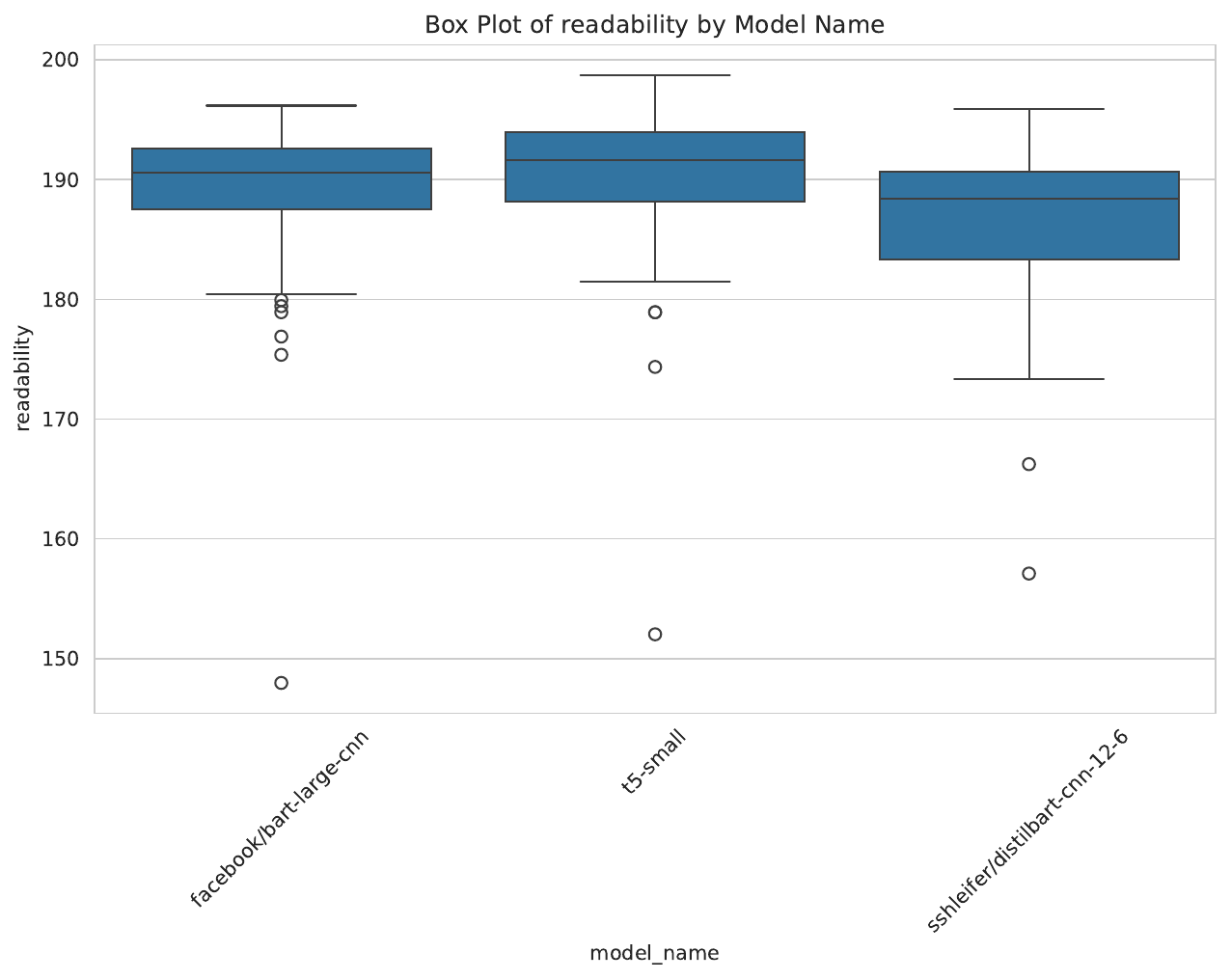}
  \caption{Readability boxplot}
  \label{fig:readability}
\end{subfigure}
\caption{Overall comparison of evaluation metrics.}
\label{fig:evaluation_metrics}
\end{figure}

Table~\ref{table:metrics_summary} provides the results when comparing the summarisation models using the evaluation metrics defined in section \ref{sec:evalmetrics}. Fig.~\ref{fig:evaluation_metrics} provides an overview of the performance and variability of each model across concepts. Among the models, BART achieves the best overall performance and generalisation. BART creates relevant and coherent summaries that maximise vocabulary diversity, and it attains this reliable performance as evidenced by low standard deviations in key metrics like diversity, relevance, and coherence. Below is a summary of the results shown in Table~\ref{table:metrics_summary}. \textbf{Diversity:} BART's diversity score is $0.806 \pm 0.058$ standard deviation, indicating its summaries incorporate the richest unique vocabulary with low variability run-to-run. \textbf{Relevance:} BART has highest relevance at $0.922 \pm 0.015$. This shows its summaries strongly preserve source semantics with minimal fluctuation. \textbf{Coverage:} All models exhibit low coverage of $\sim 0.07 \pm 0.03$ standard deviations because summaries significantly condense full text. \textbf{Coherence:} BART's scores $0.794 \pm 0.022$ coherence, meaning summary sentences interrelate accurately. This combination enables easy comprehension. \textbf{Conciseness:} The models achieve similar conciseness ($\sim 0.02 \pm 0.004$), quantifying the summary brevity. \textbf{Readability:} T5-small tops at $190.288 \pm 5.799$ in linguistic simplicity. But BART nearly matches at $189.148 \pm 5.863$.  
\vspace{-0.3in}
\begin{table}[ht]
\centering
\caption{Evaluation metrics for summarisation models}
\label{table:metrics_summary}
\begin{tabular}{lccc}
\hline
\textbf{Metric} & \textbf{BART} & \textbf{DistilBART} & \textbf{T5-small} \\
\hline
Diversity Score & 0.806 $\pm$ 0.058 & 0.770 $\pm$ 0.057 & 0.771 $\pm$ 0.059 \\
Relevance & 0.922 $\pm$ 0.015 & 0.709 $\pm$ 0.028 & 0.818 $\pm$ 0.035 \\
Coverage & 0.070 $\pm$ 0.034 & 0.072 $\pm$ 0.034 & 0.064 $\pm$ 0.032 \\
Coherence & 0.794 $\pm$ 0.022 & 0.674 $\pm$ 0.038 & 0.670 $\pm$ 0.030 \\
Conciseness & 0.021 $\pm$ 0.004 & 0.019 $\pm$ 0.003 & 0.022 $\pm$ 0.004 \\
Readability & 189.148 $\pm$ 5.863 & 186.896 $\pm$ 6.253 & 190.288 $\pm$ 5.799 \\
\hline
\end{tabular}
\end{table}

\vspace{-0.3in}
\subsection{BART's summarisation performance across concepts}
\begin{figure}[!t]
    \centering
    \begin{subfigure}{0.33\textwidth}
\includegraphics[width=\linewidth]{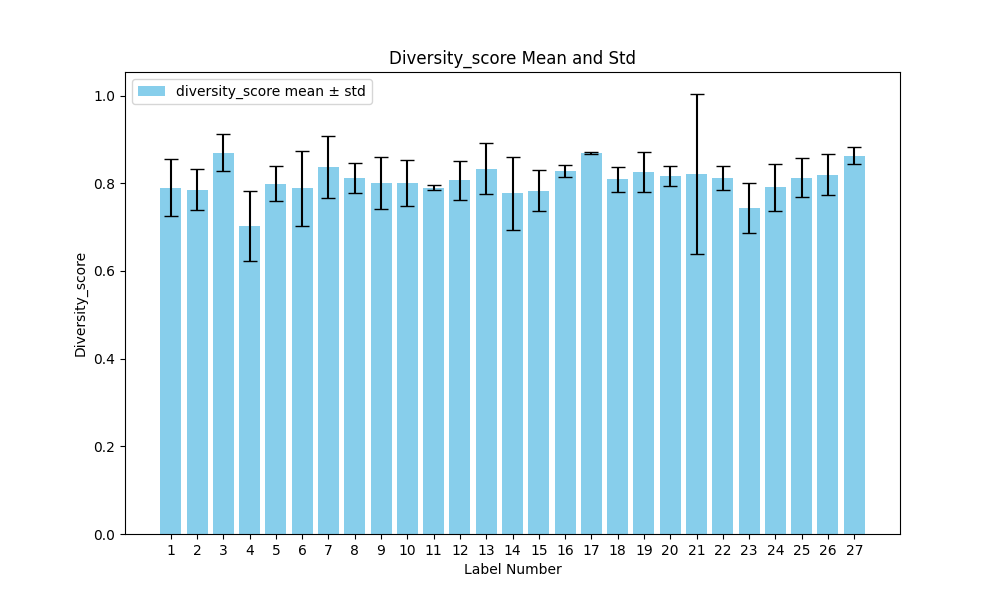}
        \caption{Diversity Score}
    \end{subfigure}%
    \hfill
    \begin{subfigure}{0.33\textwidth}
        \includegraphics[width=\linewidth]{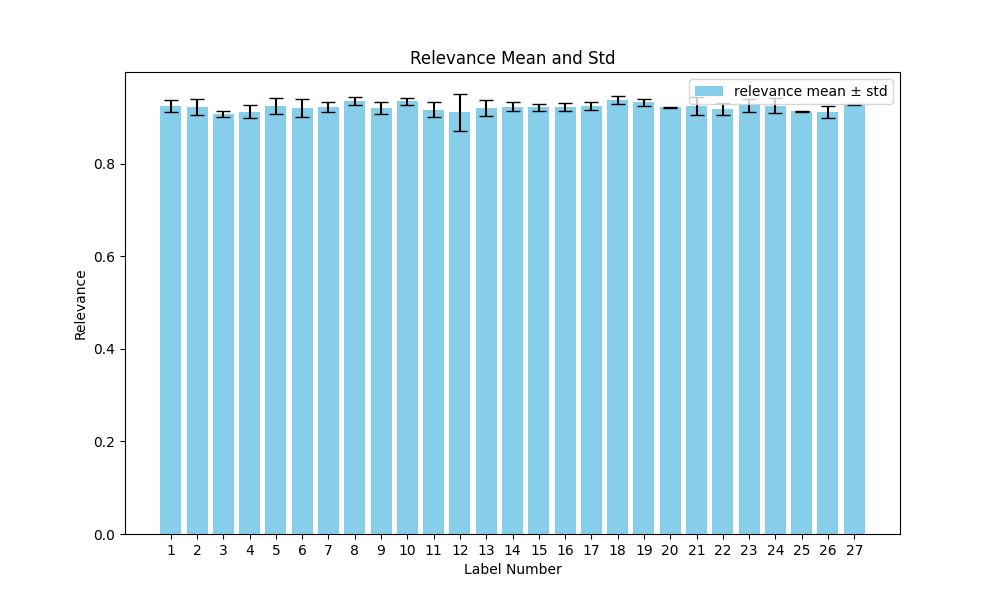}
        \caption{Relevance}
    \end{subfigure}
    \hfill
    \begin{subfigure}{0.33\textwidth}
        \includegraphics[width=\linewidth]{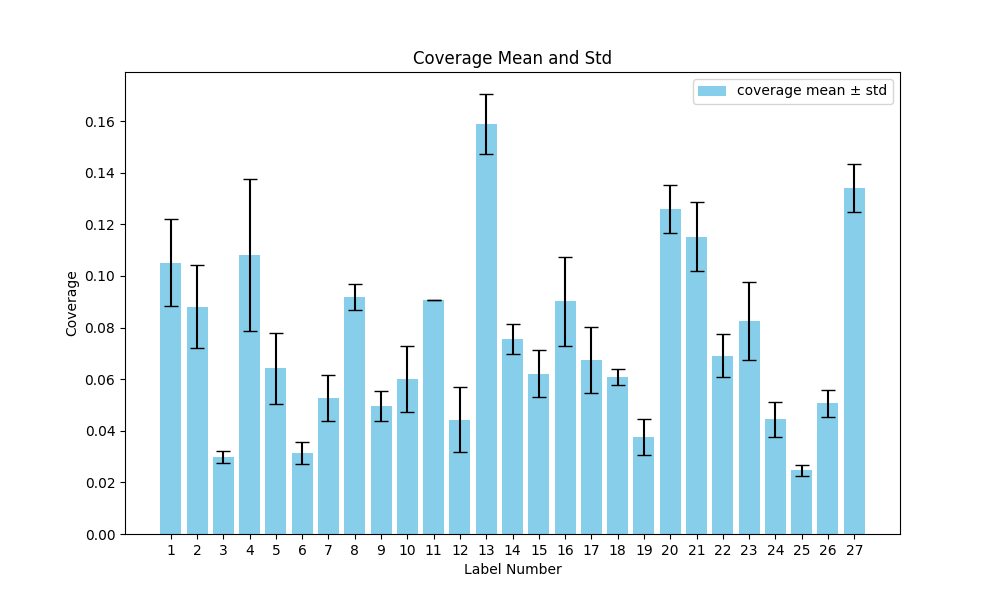}
        \caption{Coverage}
    \end{subfigure}%
    \hfill
    \begin{subfigure}{0.33\textwidth}
        \includegraphics[width=\linewidth]{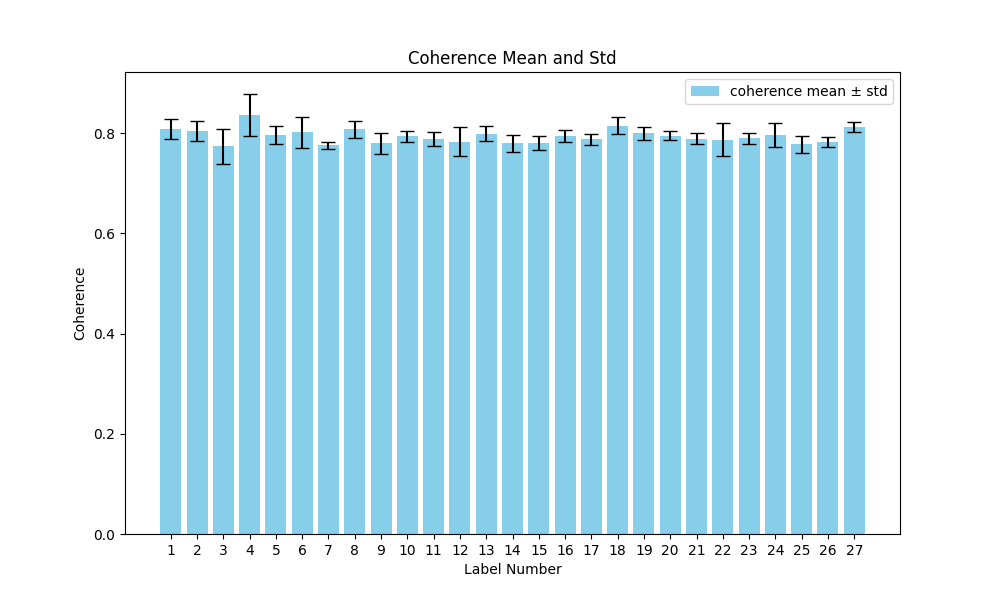}
        \caption{Coherence}
    \end{subfigure}
    \hfill
    \begin{subfigure}{0.33\textwidth}
        \includegraphics[width=\linewidth]{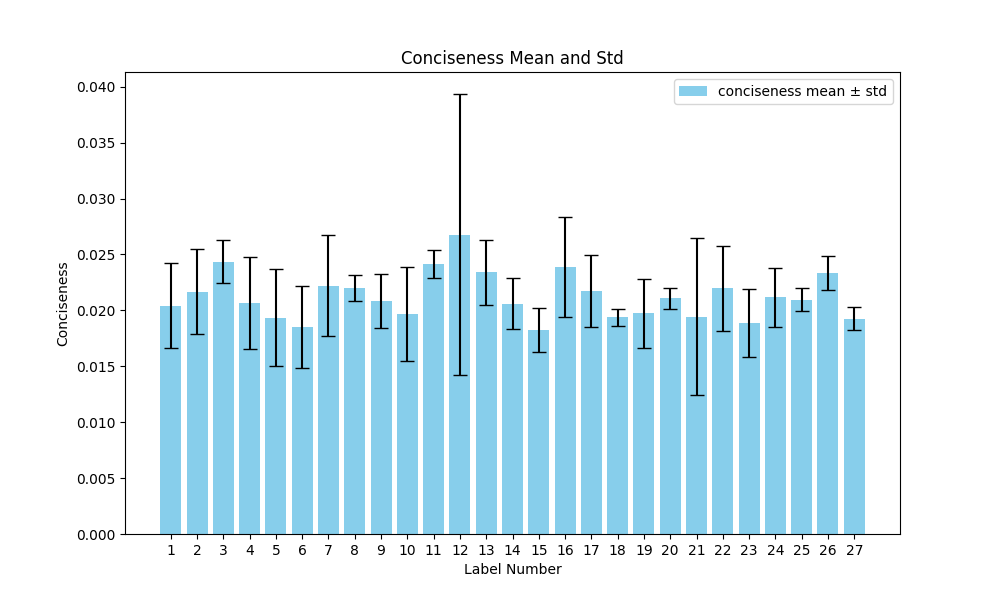}
        \caption{Conciseness}
    \end{subfigure}%
    \hfill
    \begin{subfigure}{0.33\textwidth}
        \includegraphics[width=\linewidth]{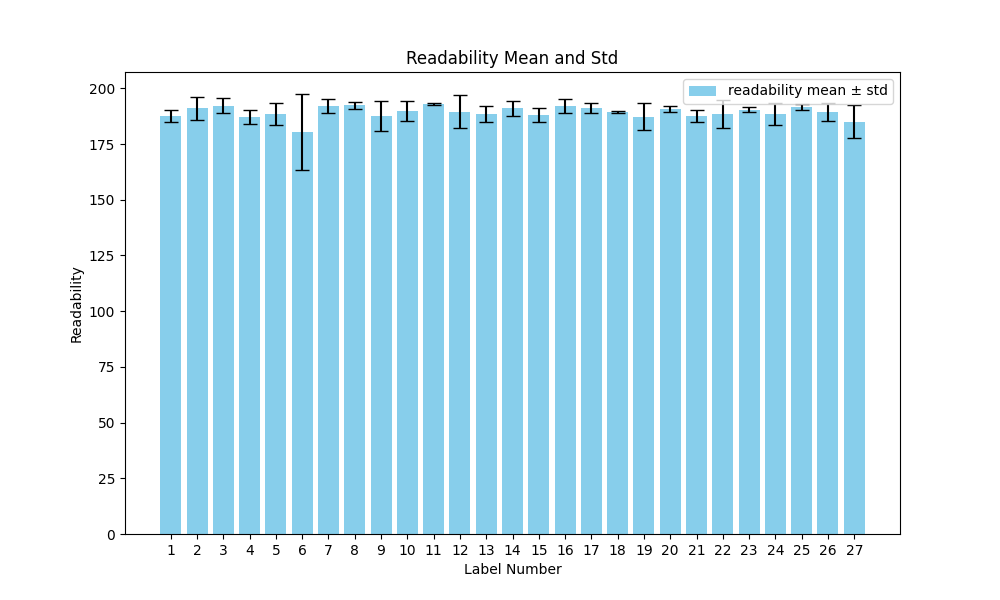}
        \caption{Readability}
    \end{subfigure}
    \caption{Summary of model evaluation metrics for BART.}\label{facebookperformance}
\end{figure}

Fig.~\ref{facebookperformance} depicts the performance of BART's summarisation model across each concept. The x-axes show the concept numbers, and these correspond to the IDs found in Table \ref{Labels}. Hence number 1 on the x-axis of Fig. \ref{facebookperformance}(a) corresponds to the Acuity concept that has ID 1 in Table \ref{Labels}.\\
The areas of strength are: \textbf{Relevance:} Across all labels, the model consistently showed high relevance scores, with means above 0.9. This indicates that the model is adept at generating content that is closely related to the given topics. Low standard deviations in relevance scores, such as 0.012 for Acuity and 0.007 for Assessment, investigation, testing, screening, suggest that the model maintains this relevance across different instances reliably. \textbf{Coherence:} The model also performed well in terms of coherence, particularly for labels such as COVID, with a mean of 0.836 and Acuity with a mean of 0.809. These scores suggest that the generated text logically flows from one sentence to the next, making it easier for readers to follow the narrative. \\The areas for improvement are: \textbf{Coverage:} The model struggled with coverage across several labels, notably so for Assessment, investigation, testing, screening, with a mean of 0.03 and a very low standard deviation, indicating consistently narrow topic coverage. This suggests that the model may not fully address all relevant aspects of a topic, possibly omitting important details. \textbf{Diversity:} While not as critical as coverage, diversity scores varied more widely across labels, with COVID showing a notably lower mean (0.702) and higher standard deviation (0.079). This indicates that the model's ability to generate varied text differs significantly across topics, which could limit its effectiveness in engaging readers with novel content or perspectives. \textbf{Readability:} The readability scores were consistently high across labels, with means around 187 to 192, suggesting the text may be complex and potentially challenging for a general audience to understand. While not a direct measure of performance like relevance or coherence, high readability scores indicate room for improvement in making the content more accessible to a broader audience.

\vspace{-0.3cm}
\subsection{Evaluating equity in summaries of investigation reports} 
\begin{figure}[h]
\centering
\includegraphics[width=0.85\linewidth]{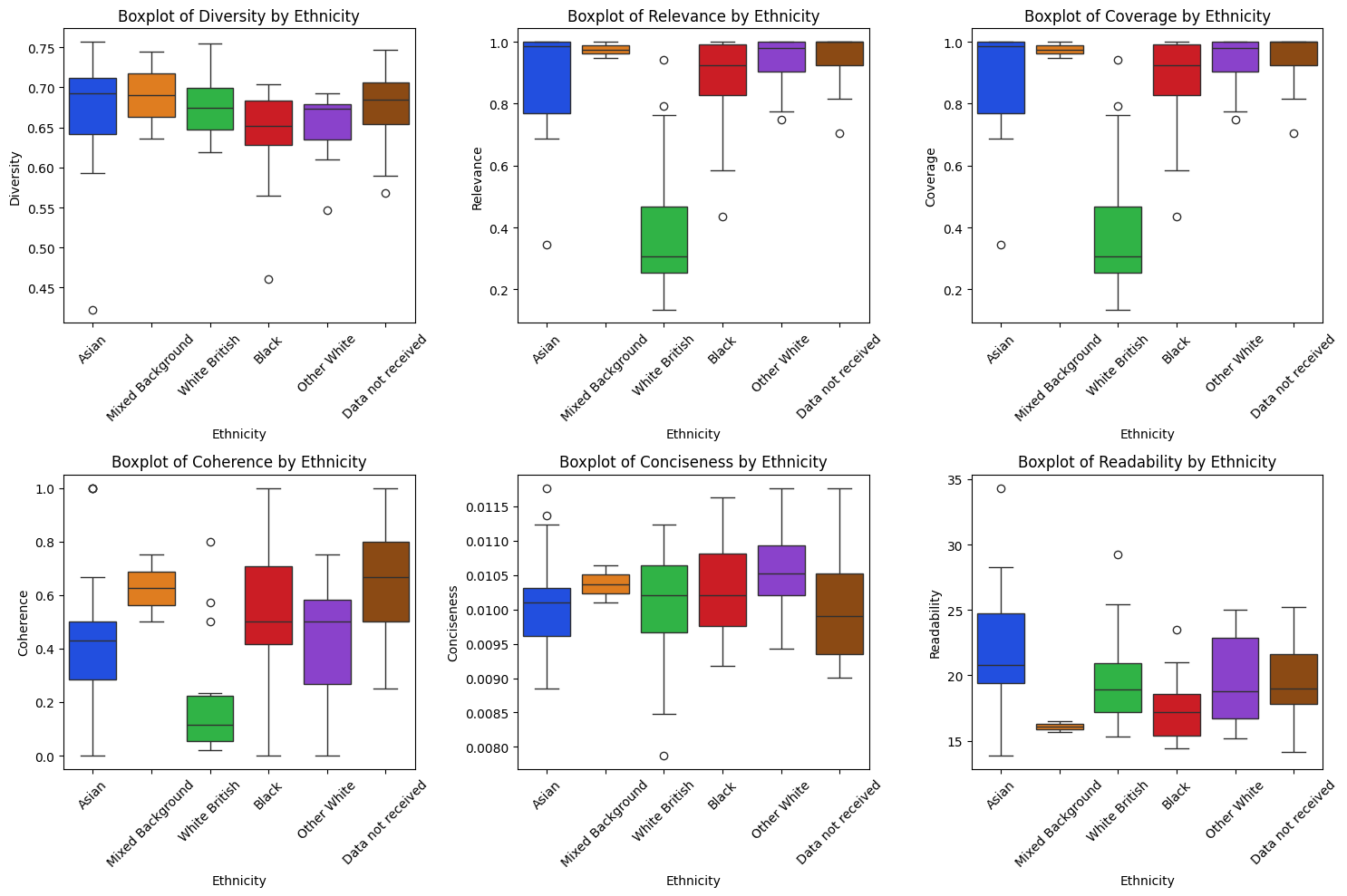}
\caption{Results per ethnic group when using BART}
\label{ethnicity}
\end{figure}

\noindent Table \ref{table:metrics_by_ethnicity} and Fig. \ref{ethnicity} present a summary of various evaluation metrics across different ethnic groups: Asian, Black, Data not received (DNR), Mixed Background (MB), Other White (OW), and White British (WB) when using BART. 

\begin{table}[]
\centering
\caption{Average and std. values of metrics by ethnicity when using BART}
\begin{tabular}{lcccccc}
\hline
\textbf{Metric} & \textbf{Asian} & \textbf{Black} & \textbf{DNR} & \textbf{MB} & \textbf{OW} & \textbf{WB} \\
\hline
\textbf{Diversity} & \(0.67 \pm 0.08\) & \(0.64 \pm 0.07\) & \(0.67 \pm 0.06\) & \(0.69 \pm 0.08\) & \(0.65 \pm 0.04\) & \(0.68 \pm 0.04\) \\
\textbf{Relevance} & \(0.89 \pm 0.18\) & \(0.87 \pm 0.17\) & \(0.95 \pm 0.09\) & \(0.97 \pm 0.04\) & \(0.93 \pm 0.10\) & \(0.40 \pm 0.22\) \\
\textbf{Coverage} & \(0.89 \pm 0.18\) & \(0.87 \pm 0.17\) & \(0.95 \pm 0.09\) & \(0.97 \pm 0.04\) & \(0.93 \pm 0.10\) & \(0.40 \pm 0.22\) \\
\textbf{Coherence} & \(0.46 \pm 0.32\) & \(0.54 \pm 0.29\) & \(0.65 \pm 0.26\) & \(0.63 \pm 0.18\) & \(0.42 \pm 0.27\) & \(0.18 \pm 0.19\) \\
\textbf{Conciseness} & \(0.01 \pm 0.00\) & \(0.01 \pm 0.00\) & \(0.01 \pm 0.00\) & \(0.01 \pm 0.00\) & \(0.01 \pm 0.00\) & \(0.01 \pm 0.00\) \\
\textbf{Readability} & \(21.54 \pm 5.12\) & \(17.53 \pm 2.54\) & \(19.23 \pm 3.37\) & \(16.08 \pm 0.59\) & \(19.54 \pm 3.63\) & \(19.62 \pm 3.58\) \\
\hline
\end{tabular}
\label{table:metrics_by_ethnicity}
\end{table}
\noindent \textbf{Diversity:} The diversity scores are relatively similar across all ethnicities, ranging from 0.64 to 0.69, with small standard deviations (0.04 to 0.08). This suggests that the BART model generates diverse outputs for all ethnicities. \textbf{Relevance and Coverage:} The relevance and coverage scores are identical for each ethnicity, indicating that these metrics are closely related. Asian, Black, DNR, MB, and OW have high scores (0.87 to 0.97) with small standard deviations (0.04 to 0.18), suggesting that the model generates relevant and comprehensive outputs for these ethnicities. However, WB has a lower score (0.40) with a larger standard deviation (0.22), indicating that the model's outputs for this ethnicity may be less relevant and comprehensive. \textbf{Coherence:} The coherence scores vary across ethnicities, with DNR and MB having the highest scores (0.65 and 0.63) and moderate standard deviations (0.26 and 0.18). Asian, Black, and OW have lower scores (0.42 to 0.54) with larger standard deviations (0.27 to 0.32), suggesting that the model's outputs for these ethnicities may be less coherent. WB has the lowest coherence score (0.18) with a small standard deviation (0.19), indicating that the model's outputs for this ethnicity are the least coherent. \textbf{Conciseness:} All ethnicities have the same conciseness score (0.01) with no standard deviation, suggesting that the model generates equally concise outputs for all ethnicities. \textbf{Readability:} The readability scores range from 16.08 to 21.54, with standard deviations ranging from 0.59 to 5.12. MB has the lowest readability score (16.08) with the smallest standard deviation (0.59), indicating that the model's outputs for this ethnicity are the most readable and consistent. Asian has the highest readability score (21.54) with the largest standard deviation (5.12), suggesting that the model's outputs for this ethnicity are the least readable and have the most variability. In summary, the BART model generates diverse, relevant, and concise outputs for most ethnicities, with some variations in coherence and readability. The model's performance appears to be the weakest for the WB ethnicity, with lower relevance, coverage, and coherence scores. These low scores could be attributed to the sample size and variability of the WB sentences. While the WB group has the most sentences, this does not guarantee better performance. A larger dataset can introduce more variability, making it harder for the model to learn consistent patterns.
Table \ref{samplesummaries} presents two sample summaries (from reports of different ethnic groups) generated by the BART summarisation model. 
\begin{table}[!h]
\centering \small
\caption{Sample summaries from multiple reports for two ethnic groups. Ethnicities cannot be disclosed. Each summary is generated from multiple reports.}
\begin{tabular}{p{1in}|p{4in}}
\hline
Concept  & Summary\\
\hline
Communication & \textbf{Staff not heard.} Staff voiced their concerns about this decision to the senior obstetrician. They were left feeling that their concerns had not been heard. There was no formal debriefing afterwards, which staff would have valued. The opportunity to share reflections and learning was not completed with all staff involved. The incremental delays caused by finding and allocating staff to the concurrent theatre cases, and communication breakdown within the team further impacted on the DDI.\\\hline
Acuity  & \textbf{Reviews with seniors did not occur. }The ambulance Trust was experiencing high volumes of 999 calls at the time of a Mother's call. Due to the high acuity on the labour ward the initial decisions were not discussed with the senior clinician. A senior face to face review did not occur until 16:05 hours, 2 hours and 50 minutes after the initial recognition of abnormalities of the Baby's heart rate. The abnormal CTG trace from the IOL suite was not reviewed by the senior obstetrician.\\ 
\hline
\end{tabular} 
\label{samplesummaries}
\end{table} 
\vspace{-0.1in}
\section{Ethical risks of abstractive summarisation models}
\begin{itemize}
\item \textbf{Risk of information hallucination and bias amplification:} Abstractive summarisation models like BART, T5, and DistilBART risk generating content not found in the original data and amplifying biases from their training data. \textbf{Mitigation:} Diversifying training datasets, implementing debiasing techniques, and cross-referencing with original texts. Regular bias audits and integrating ethical guidelines directly into the training process can enhance accuracy, transparency, and fairness.
\item \textbf{Risk of inadequate control over content:} Unlike extractive methods, abstractive models generate summaries based on learnt patterns, which can result in misrepresentation or underrepresentation of certain groups or incidents. \textbf{Mitigation:} Developing more sophisticated models with embedded ethical and fairness constraints, and employing precise source mapping mechanisms like sentence IDs, can strengthen control and ensure verifiability and trustworthiness of the content generated.
\item \textbf{Risk in processing sensitive data:} Utilising online LLMs for sensitive data can pose risks of data breaches and privacy violations. \textbf{Mitigation:} Switching to offline LLMs lessens these risks but may reduce performance, as it depends on local computing resources. Organisations need to balance security against performance to find an appropriate balance.
\item \textbf{Sample Size and Variability:} Large datasets, whilst rich in information, present challenges due to variability and the potential inclusion of low-quality data. This complexity hinders a model's ability to learn consistent patterns, adversely affecting summarisation metrics. \textbf{Mitigation:} Explore robust data preprocessing techniques to enhance data quality. Utilising machine learning algorithms customised for high variability will enhance consistency and equity in summarisation outcomes across diverse groups.
\end{itemize}

\section{Conclusion}
The I-SIRch:CS framework automates the analysis and summarisation of textual data in maternity incident reports, holding significant potential for uncovering critical insights and contributing factors to preventable harm. Future work will prioritise enhancing the framework's traceability capabilities by implementing mechanisms to provide clear links between generated summaries and original reports, allowing for easy verification of accuracy and context. Explainable AI techniques will also be explored to offer insights into how models generate summaries. A key limitation of the I-SIRch:CS framework is the lack of Patient and Public Involvement (PPI) in assessing its outputs. PPI is crucial as it ensures solutions are relevant to patient needs and experiences, potentially improving the framework's summaries and their real-world applicability. Engaging with patients and the public could also increase the model's transparency and trustworthiness. Future development will aim to integrate PPI feedback, enhancing the framework's effectiveness and its contributions to patient safety and care quality.

\begin{credits}
\subsubsection{\ackname} This report is independent research funded by NHSX and The Health Foundation and it is managed by the National Institute for Health Research (AI\_HI200006). The views expressed in this publication are those of the author(s) and not necessarily those of the NHSX, The Health Foundation, National Institute for Health Research, or the Department of Health and Social Care.

\subsubsection{\discintname}
The authors have no competing interests to declare that are
relevant to the content of this article. 
\end{credits}

 \bibliographystyle{splncs04}
\bibliography{references}

\end{document}